\documentclass{article}

\usepackage{arxiv}
\usepackage[utf8]{inputenc} 
\usepackage[T1]{fontenc}    
\usepackage{hyperref}       
\hypersetup{colorlinks,allcolors=black}
\usepackage{url}            
\usepackage{booktabs}       
\usepackage{amsfonts}       
\usepackage{nicefrac}       
\usepackage{microtype}      
\usepackage{lipsum}		
\usepackage{graphicx}
\usepackage[square,sort,comma,numbers]{natbib}
\usepackage{doi}
\usepackage{array}

\title{\textsc{DSSE}: a Drone Swarm Search Environment}

\author{
        {Manuel Castanares} \\
        {Insper. São Paulo, Brazil}\\
	{manuelc@al.insper.edu.br} \\
	\And
        {Luis F. S. Carrete} \\
        {Insper. São Paulo, Brazil}\\        
	{luisfsc@al.insper.edu.br} \\
	\And
        {Enrico F. Damiani} \\
        {Insper. São Paulo, Brazil}\\
	{enricofd@al.insper.edu.br} \\
	\And
        {Leonardo D. M. de Abreu} \\
        {Insper. São Paulo, Brazil}\\
	{leonardodma@al.insper.edu.br} \\
	\And
        {José Fernando B. Brancalion} \\
        {Embraer. São José dos Campos, Brazil}\\
	{jose.brancalion@embraer.com.br} \\
	\And
        {Fabrício J. Barth} \\
        {Insper. São Paulo, Brazil}\\
	{fabriciojb@insper.edu.br}
}

\hypersetup{
pdftitle={\textsc{DSSE}: a Drone Swarm Search Environment},
pdfsubject={},
pdfauthor={},
pdfkeywords={Reinforcement Learning, Simulation, Multi Agent Systems, Maritime search and rescue},
}

\begin{document}
\maketitle

https://capstone-insper.github.io/drone-swarm-search/

\begin{abstract}
The Drone Swarm Search project is an environment, based on \textsc{PettingZoo}, that is to be used in conjunction with multi-agent (or single-agent) reinforcement learning algorithms. It is an environment in which the agents (drones), have to find the targets (shipwrecked people). The agents do not know the position of the target and do not receive rewards related to their own distance to the target(s). However, the agents receive the probabilities of the target(s) being in a certain cell of the map. The aim of this project is to aid in the study of reinforcement learning algorithms that require dynamic probabilities as inputs. A peer-reviewed paper describing version 2 of this software has been published in \cite{falcao_dsse_2024}. 
\end{abstract}


\section{Introduction}

Every year, vast bodies of water worldwide claim numerous missing individuals. According to the World Health Organization (WHO), there are an estimated 236,000 annual drowning deaths worldwide, making it the third leading cause of unintentional injury death worldwide and accounting for 7\% of all injury-related deaths \citep{drone2021}. With over 71\% of the earth's surface covered by oceans, according to the U.S. Geological Survey (USGS) \citep{survey2019}, finding these missing individuals is no easy task, due to the complexity of oceanic environments and the vastness of the search areas. However, drone swarms have emerged as a promising tool for searching for missing individuals.

The use of drones in rescue operations has resulted in successfully saving 940 people while being utilized in 551 rescue incidents so far \citep{drone2023}. The capacity of drones to reach difficult terrain and inaccessible areas, as well as their ability to capture real-time images and videos, has proved to be helpful in search and rescue missions.

The accuracy of search and rescue missions is believed to be significantly increased by the incorporation of Artificial Intelligence (AI) technology \citep{eliack2022}, as it can leverage probabilistic models based on the ocean’s behaviors, as well as the last known location of the people being rescued.

Several solutions have been proposed in the last years to solve this problem \citep{alotaibi2019,ai2021,schuldt2017,wang22}, in special, using reinforcement learning algorithms. The reinforcement algorithm does not work alone, it needs an environment to guide it. Because of that, all of the articles cited below have created their own environment to recreate the real-world scenario in a way that the algorithm could understand what was happening and what has to be done. 

However, all the papers that claimed to use personalized environments did not make it available to the public, and all of the tools were developed for internal use only. The fact of not publishing those tools as public ones

The fact of not publishing those tools as public resources can be seen as a limitation in terms of reproducibility, transparency, and collaboration in the field of reinforcement learning. It restricts the ability of other researchers and practitioners to build upon or validate the work conducted in those papers. For this reason, the goal of this paper is to provide a tool for everyone so that interested parties can search for better solutions to the problem of finding shipwrecked people using a drone swarm.

This preprint describes the initial design and release of the tool. An updated software paper with expanded functionality and additional contributors was later published in \cite{falcao_dsse_2024}.

The rest of the paper is organized as follows. Next section presents the simplications adopted in order to build the environment. Section \ref{sec:understanding} presents all theory related to the development of the probability matrix. Section \ref{sec:movement} describes the target's movement algorithm. Sections \ref{sec:rewards} and \ref{sec:implementation} describe the reward function and the implementation of the environment as a python library, respectaly. Section \ref{sec:real} gives details about the relation between this tool and real world. Finally, section \ref{sec:conclusion} offers conclusions and directions for future work.

\section{Adopted simplications}

To achieve what was proposed, a simulation of the real-world situation had to be created, but since the real world is incredibly complex a few premises had to be set to simplify the overall scenario. First of all, there will only be one shipwrecked person, since adding multiple people would increase the complexity of the algorithm. It is considered that once the drone is over the person, and executes the action of search, he will identify the person. How he identifies the person is not considered, since this is only a simulated environment. The drone will be able to move only in the area delimited by a grid that covers where the shipwrecked person is located. The drone can only execute five different actions, moving up, down, left, right, and searching. The search was defined as an action because, in this scenario, the drone will be flying at a high altitude so that it can visualize a bigger area, once it identifies a possible target it will descend and verify that it is in fact a person, therefore the search action was included to represent this process. The drone also does not move diagonally to simplify the model. The environment will not simulate the wind or any natural disaster which may affect the drone's flight.
	
\begin{figure}[htbp]
	\centering
	\includegraphics[scale=0.25]{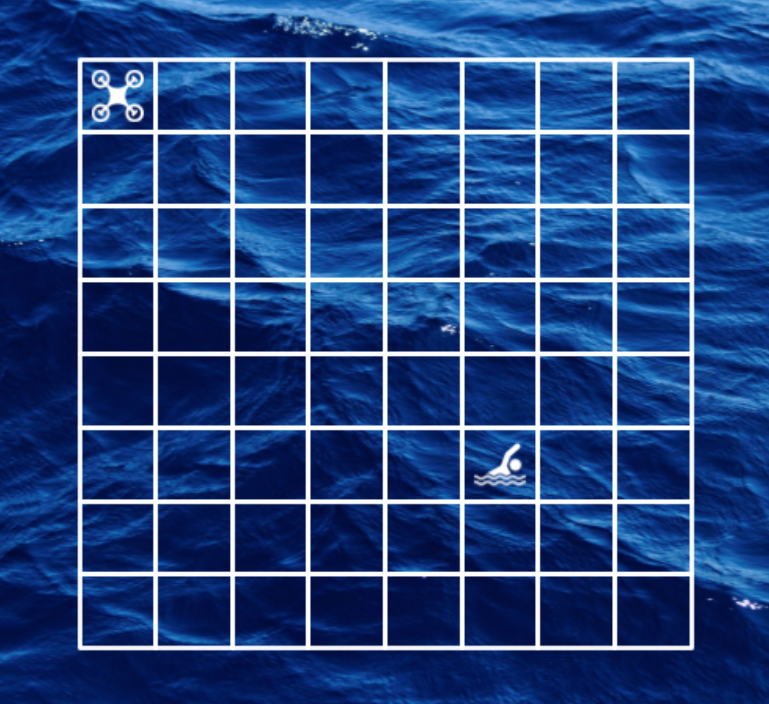}
	\caption{Map representation}
	\label{fig:map}
\end{figure}

The drones have a restriction, their battery life. They can only do a certain number of steps before their battery ends. The person's movement in the ocean will also not be defined by complex ocean modeling but by a simple vector that will force the person to drift away over time. Finally, the drone will be placed in a grid similar to the one below (image \ref{fig:map}), where each cell has a probability, representing the chances of the shipwrecked person being located in that area.

\section{Understanding the probability matrix}
\label{sec:understanding}

Based on previous studies \citep{schuldt2017,ai2021} a probability matrix will be created to demonstrate the chances of the shipwrecked person being in a given cell. The matrix has the same dimensions as the position matrix, and it is the primary piece of information used by the agent. Researchers with similar areas of study \citep{schuldt2017,ai2021} used multiple metrics in order to define the values of the matrix. For example, the wind and flow of the ocean can greatly impact the trajectory of a shipwrecked person. This type of data can vary depending on the place, day, and time. However, since the modeling of the ocean is not a priority of the project, a directional vector will act as the ocean’s current, which will subsequently drag the shipwrecked person to different places on the map. This will, therefore, change the value of each cell in the probability matrix. Said directional vector along with the initial position of the shipwrecked person are inputs that can be defined by the user. This allows the simulation of a scenario in which the user has knowledge of the ocean’s current, along with the shipwrecked person’s last known localization.
	
\begin{figure}[htbp]
	\centering
	\includegraphics[scale=0.3]{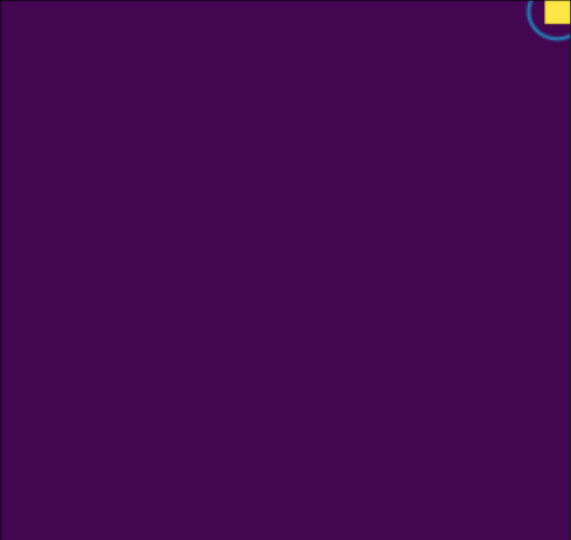}
	\caption{Initial state of a probability matrix (color represents probability)}
	\label{fig:initial_state}
\end{figure}

Using the directional vector and the initial position of the shipwrecked person, the probability matrix can be created. In the first state, a probability of one hundred percent is placed in the cell where the person was last seen (image \ref{fig:initial_state}). As time progresses, the supposed position of the person moves according to the directional vector. Once the assumed position of the person moves, according to the vector, the probabilities are distributed around this new cell. Additionally, a circumference is placed around the cell in which the person is assumed to be. This circumference dictates the area in which the person could be (probability greater than zero). The radius of this circle is increased as time goes by, to represent the growing uncertainty of the position of the shipwrecked person. All of the cells that are inside of the circle receive a probability, which is calculated using the bi-dimensional Gaussian function (equation \ref{eq:gaussian}).

\begin{equation}
    f(x,y) = A \times e^{-(\frac{(x-x_{o})^{2}}{2\sigma x^{2}}+\frac{(y-y_{0})^{2}}{2 \sigma y^{2}})}
    \label{eq:gaussian}
\end{equation}

where $A$ is the amplitude, $x_{0}$ and $y_{0}$ is the supposed position of a person, and $\sigma x$, $\sigma y$ define how the function will be stretched along the matrix. Except for the supposed position of the shipwrecked person, all of these parameters are inputs. Furthermore, this formula will create a Gaussian distribution where the maximum is $A$. In order to transform the values into probabilities, the value of each cell is divided by the sum of all of the values returned by the function.

\begin{figure}[htbp]
	\centering
	\includegraphics[scale=0.3]{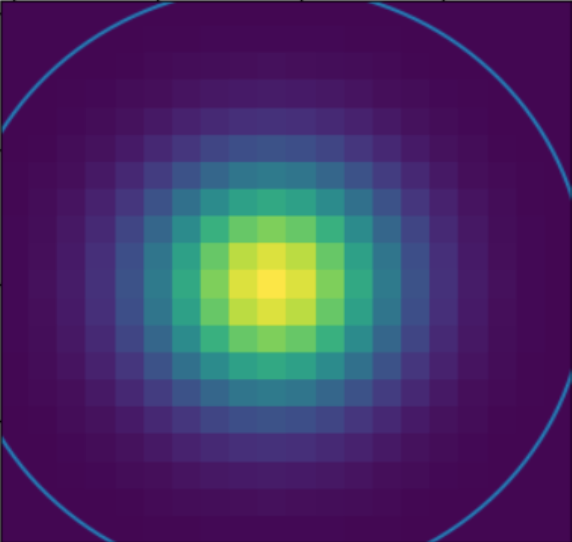}
	\caption{Probability matrix after the time (color intensity represents higher probabilities)}
	\label{fig:after_steps}
\end{figure}

As time transpires, the probability matrix will gradually change, because of the movement of the current, as well as the increase in uncertainty relating to the position of the shipwrecked person. This eventually creates a probability matrix in which multiple cells contain probabilities (image \ref{fig:after_steps}). The matrix is then used to determine the proper position of the shipwrecked person, which will not necessarily be in the cell with the highest probability.

\section{Understanding target's movement}
\label{sec:movement}

Since the goal of this library is not to simulate in depth the ocean's movements or the person's movement in the ocean, the person's movement in the grid will be created using the probability matrix which is described above. In the simulation, the person will start in a cell chosen by the user where the probability of the person being there is 100\%. In the next step, the probability will disperse, as it was described above. Considering the dispersed probability matrix the person will look at all the adjacent cell's probabilities and will make a decision either to move or to stay in its current spot, this decision is based on the probability of the adjacent cells. Therefore it is safe to assume that most times the person will choose to go to the highest probability cell making his movement follow the high probability area throughout the simulation. 
	
This movement strategy was adapted to simulate the target's decision-making, when searching for a person in the ocean, it is doubtful they will stay in the same place, they will constantly be trying to make decisions to survive, meaning, they would most likely move around. Although in a real situation, a shipwrecked person may not move as fast as the target in the simulation, the movement is also designed to simulate the uncertainty of a person being in a cell. Even though the person may not be in a high-probability cell, the agent still must search the cell, because the person will most probably be located in one of the other high-probability cells.

\section{Environment rewards}
\label{sec:rewards}

The reward is a simple concept where you will penalize the agent if it does something that it is not supposed to do and reward it in case it does something that leads it to its goal. Any reinforcement algorithm works in such a way that it will always try to maximize the agent's rewards, so if the agent does something it is not supposed to do, it will receive a massive negative reward so it learns to not do it again.

In this environment the agent receives a reward of $1$ per action by default, that is because the drone needs to be incentivized to walk and explore the grid. The drone (agent) will receive a reward of $-100000$ in case it leaves the grid. This is because in the early experiments when the reward for leaving the grid was $-1000$, the agent would learn that leaving the grid instantly would give a better reward than searching and not finding the target since if he left the grid the reward would be only $-1000$ and if he searched it and not found would be about $-2000$. Therefore the reward for leaving the grid was raised to $-100000$ so that the agent quickly learns to not leave the grid.

The agent will receive a reward of $-1000$ if it does not find the target. This is because the agent must be penalized for not finding the target but it can’t be as big as leaving the grid since the agent must still be incentivized to look for the target. The agent will receive a reward of $-2000$ in case of collision. This is because the reward needs to be lower than the case in which the agent does not find the target, otherwise, the agents would learn to crash so that they don't get a worse reward.
	
In case the agent searches, it will receive a reward according to the probability of the cell, so if the drone searches in a cell with a probability of 80\% the drone will receive a reward equal to $80$, this is because the agent needs to learn that it is better to search in higher probability cells rather than waste time searching in the lower probability areas.
	
Finally, if the drone finds the target it will receive a reward according to the equation \ref{eq:target}. This is because the agent needs to be incentivized to find the target in the fewest moves possible. So if the total timestep is $500$ and it finds the target in the timestep $480$, it will receive a reward of $200$. Still, if the agent finds in $100$ steps, it will receive a reward of $4000$, greatly incentivizing him to find the target in the quickest way possible.  

\begin{equation}
r = 10000 + 10000 * (\frac{1-timestep}{timestep\_limit})    
\label{eq:target}
\end{equation}

The variable $timestep$ represents the number count of the action that is taking place. For example, if an agent executed $50$ actions, the timestep is equal to $50$. $timestep\_limit$ is the amount of actions that an agent can do in an episode. The table \ref{table:reward} summarizes the environment rewards. 

\begin{table}
\centering
\begin{tabular}{|m{4em}|m{8em}|m{16em}|m{14em}|} 
 \hline
 \textbf{Action} & \textbf{Context} & \textbf{Description} & \textbf{Reward} \\ 
 \hline\hline
 Move & Default & Default action of going up, down, left or right & $1$ \\
 \hline
Move 
& Leave the grid
& When moving, leave the grid.
& $-100000$ \\
\hline
Move
& Collide with other agents
& When moving, go to a cell occupied by other agents.
& $-100000$ \\
\hline
Search
& Cell containing a probability smaller than 1\%
& When performing a search, the cell has a probability of less than $1$ of containing the shipwrecked person.
& $-100$ \\
\hline
Search
& Cell containing a  probability equal or bigger than 1\%
& When performing a search, the cell has a probability of $1$ or more of containing the shipwrecked person.
& Probability of the cell being searched * 10000 \\
\hline
Search
& Cell containing the shipwrecked person
& When performing a search, the cell contains the shipwrecked person.
& $10000 + 10000 * (\frac{1-timestep}{timestep\_limit})$ \\
\hline
Any of the above
& Make more actions than allowed by the timestep limit.
& More actions than the timestep limit have already occurred by the time a specific action is performed.
& $-100000$ \\ 
 \hline
\end{tabular}
\vspace{0.3cm}
\caption{Reward summary}
\label{table:reward}
\end{table}

\section{Environment implementation}
\label{sec:implementation}

The implementation of Reinforcement Learning algorithms implies the necessity of an environment, which the agents can act upon. This environment also provides multiple mechanics that are crucial for reinforcement learning. For example, the reward that each agent receives is determined by the environment. Moreover, the actions and their consequences are all embedded inside this structure. All of these aspects, and more, are necessary for the development of any reinforcement learning algorithm. Because of such dependency, it is important to maintain a certain structure and standard that allows these algorithms to be implemented in different environments, with small adaptations. The step function can be used as an example to understand the dynamic previously explained. This function is used whenever the algorithm wants the agents to perform an action. For example, in this project, the step function is responsible for moving the drones and performing the search action, whenever the algorithm wants them to. In addition, after the action is performed, its reward is calculated and returned by the same function, along with other important information. The inputs and outputs of this function, and their respective data structures (lists, dictionaries, variables, etc) all have to be in line with a certain norm. This way, when a reinforcement learning algorithm is implemented on top of this environment, the programmer can be certain that the step function’s inputs and outputs will have the same structure as other environments. The same can be said for the other functions that these algorithms require.

For this library, it was decided that the environment would follow the norms of a project called \textsc{PettingZoo} \citep{farama2023}, which makes available an array of different environments. \textsc{PettingZoo} was created by \textsc{OpenAI} as one of several tools developed to conduct artificial intelligence research along with Gymnasium, Minari, and several others. This library does not include training algorithms, as its sole purpose is to deliver specific environments. \textsc{PettingZoo} contains environments for multiple Atari games, such as Space Invaders, Pong, Mario Bros, and many more. This way, reinforcement learning algorithms can be created on top of these video games. These environments can be understood as a shell that can fit many different training algorithms so that users can study and improve different training algorithms without having to worry about recreating the environments.

Finally, a python package\footnote{\href{https://pypi.org/project/DSSE/}{https://pypi.org/project/DSSE/}} with this environment was created, with the intention of making it available for future studies. The source code for this package is also publicly available on GitHub\footnote{\href{https://capstone-insper.github.io/drone-swarm-search/}{https://capstone-insper.github.io/drone-swarm-search/}}. This repository contains a detailed documentation of the environment, including installation instructions, thorough descriptions of functions and variables, and an example of how this environment is to be used. 

\section{Environment and the real world}
\label{sec:real}

For the environment to be useful in a real scenario, the dimensions of the environment need to be determined in relation to the real world. For example, if the environment were to be used in a real-life scenario, how would the grid size be defined? For that, two pieces of information will be needed, the search zone size and the cell size. The search zone size is an independent variable that will change with every scenario, but most importantly, the cell size must be defined as a constant. Given that drones are only allowed to fly at about $120$ meters \citep{leslie2022}, because of aerial space interference, and considering the Wiris ProSc camera \citep{camera2023}, which is a camera used for environmental research, archeological and geological research, and so on, the drone’s field of view will be about $16900m^{2}$ at an altitude of $120m$. 

The environment considers that the cell is defined by what the drone can view, so whenever the drone is in a cell it can scan the whole area of the cell for the target. Therefore it is safe to assume that the size of each cell will also be $130m \times 130m$ considering the camera and altitude above. Thus in a real-world case where the search zone is $1km \times 1km$, the environment will need to be created with a grid size equal to 8, so that the grid in the simulation represents an area of $1.04km \times 1.04km$. It is important to note that this cell size is defined by the camera and altitude chosen, changing these parameters will change the cell size.

\section{Conclusion and future work}
\label{sec:conclusion}

The environment, published as a python package\footnote{\href{https://pypi.org/project/DSSE/}{https://pypi.org/project/DSSE/}}, was designed with the intention of allowing external researchers to utilize and modify it as needed. This open approach encourages others to build upon the project, potentially achieving even more remarkable results than those demonstrated here. By engaging a wider community, the utilization of this environment has the potential to drive further improvements, thereby influencing future algorithmic advancements.

One of the biggest limitations of this environment is the fact that the shipwrecked person will not leave the grid if it reaches its edge, but will simply move around in the corner until the episode is complete. Second, the drones' actions are discrete, which does not represent the real world, where the drone is free to move in any direction using a continuous space. Finally, there is also a limitation with the ocean’s simulation and the target’s movements. Although it was sufficient for the first delivery, it may not be for a real-life situation, so it would be interesting to add a more sophisticated way to calculate the probability matrix as well as a more complex simulation for the target's movement.

\bibliographystyle{plain}
\bibliography{dssepaper}  
\end{document}